# Multilevel Thresholding for Image Segmentation through a Fast Statistical Recursive Algorithm

S. Arora [a], J. Acharya [b], A. Verma [c], Prasanta K. Panigrahi [c,1]

[a] *Dhirubhai Ambani Institute of Information and Communication Technology, Gandhinagar, 382 009, India*

[b] *Indian Institute of Technology, Kharagpur, 721 302, India*

[c] *Physical Research Laboratory, Navrangpura, Ahmedabad, 380 009, India*

**Abstract**

A novel algorithm is proposed for segmenting an image into multiple levels using its mean and variance. Starting from the extreme pixel values at both ends of the histogram plot, the algorithm is applied recursively on sub-ranges computed from the previous step, so as to find a threshold level and a new sub-range for the next step, until no significant improvement in image quality can be achieved. The method makes use of the fact that a number of distributions tend towards Dirac delta function, peaking at the mean, in the limiting condition of vanishing variance. The procedure naturally provides for variable size segmentation with bigger blocks near the extreme pixel values and finer divisions around the mean or other chosen value for better visualization. Experiments on a variety of images show that the new algorithm effectively segments the image in computationally very less time.

## 1 Introduction

Thresholding is an important technique for image segmentation. Because the segmented image obtained from thresholding has the advantage of smaller storage space, fast processing speed and ease in manipulation, compared with a gray level image containing 256 levels, thresholding techniques have drawn a lot of attention during the last few years. The aim of an effective segmentation is to separate objects from the background and to differentiate pixels having nearby values for improving the contrast. In many applications of image processing, image regions are expected to have homogeneous characteristics (e.g., gray level, or color), indicating that they belong to the same object or are facets of an object, implying the possibility of effective segmentation.

[1] E-mail: prasanta@prl.res.in

Thresholding techniques can be divided into bi-level and multi-level category, depending on number of image segments. In bi-level thresholding, image is segmented into two different regions. The pixels with gray values greater than a certain value T are classified as object pixels, and the others with gray values lesser than T are classified as background pixels. Several methods have been proposed to binarize an image, Sezgin and Sankur (2004). Otsu's method (1979) chooses optimal thresholds by maximizing the between class variance. Sahoo et al. (1988) found that in global thresholding, Otsu's method is one of the better threshold selection methods for general real world images with regard to uniformity and shape measures. However, inefficient formulation of between class variance makes the method very time consuming. Abutaleb (1989) used two-dimensional entropy to calculate the threshold. In Pun's method (1980), as modified by Kapur (1985), the picture threshold is found by maximizing the entropy of the histogram of gray levels of the resulting classes. Wang et al. (2002) proposed an image thresholding approach based on the index of nonfuzziness maximization of 2D grayscale histogram. Kittler and Illingworth (1986) suggested a minimum error thresholding method. Niblack's method (1986) is a local approach which builds a threshold surface, based on the local mean, m, and local standard deviation, s, computed in a small neighborhood of each pixel in the form of T = m + k s, where k is a negative constant. This algorithm, however, produces a large amount of binarization noise in those areas that contain no text objects. Wu and Amin (2003) use a multi stage thresholding, first at global level, and then proceed locally over the image. Binarization for non-uniformly illuminated document images has been considered by Feng and Tan (2004).

Multilevel thresholding is a process that segments a gray-level image into several distinct regions. This technique determines more than one threshold for the given image and segments the image into certain brightness regions, which correspond to one background and several objects. The method works very well for objects with colored or complex backgrounds, on which bi-level thresholding fails to produce satisfactory results. Reddi et al. (1984) proposed an iterative form of Otsu's method, so as to generalize it to multilevel thresholding. Ridler and Calward algorithm (1978) uses an iterative clustering approach. An initial estimate of the threshold is made (e.g., mean image intensity); pixels above and below are assigned to the white and black classes respectively. The threshold is then iteratively re-estimated as the mean of two class means. The most difficult task is to determine the appropriate number of thresholds automatically. Unfortunately, many thresholding algorithms are not able to automatically determine the required number of thresholds, as has been noted by Whatmough (1991). Chang (1997) uses a lowpass/highpass filter repeatedly to adjust (decrease/increase) the number of peaks or valleys to a desired number of classes and then the valleys in the filtered histogram are used as thresholds. Boukharouba et al. (1985) define the zeros of a curvature function as multithreshold values by using a distribution function. Papamarkos and Gatos (1994) specify the multithreshold values as the global minima of the rational functions which approximate the histogram segments by using hill clustering technique to determine the peak locations of image histogram. Huang et al. (2005) proposed a multi-level thresholding for unevenly lighted image using Lorentz information measure. Tseng et al. (1993) used an automatic thresholding method based on aspect of human visual system for edge detection and segmentation.

Keeping in mind human visual perception, extreme pixel values need not be finely quantized. By suitable coarse graining these can be progressively removed from the rest of the pixel values, which need to be finely segmented. A recursive implementation yields a non-uniform segmentation which naturally allows finer quantization around mean. This procedure zooms in to the mean in a manner similar to the approach of a variety of distributions towards Dirac delta function:

$$\lim_{\sigma \to 0} f(x) = \lim_{\sigma \to 0} \frac{1}{\sigma\sqrt{2\pi}} \, exp\left[-\frac{(x-\mu)^2}{2\sigma^2}\right] = \delta(x-\mu)$$

## 2  Approach

In the present approach, we use mean and the variance of the image to find optimum thresholds for segmenting the image into multiple levels. The algorithm is applied recursively on sub-ranges computed from the previous step so as to find a threshold and a new sub-range for the next step. The following points have been considered while designing the proposed algorithm:

1) A large class of images have histograms having high intensity values for pixels near a certain value (generally the mean), or they have many structures at intensity values near the mean and less number of structures farther from mean. A rough estimate of such a histogram is a Gaussian distribution.

2) The human eye is not very sensitive to the features present at both the extreme pixel intensity values, but is sensitive to distinguish features present at the mid-range values of intensities. Hence, it is useful to concentrate about the middle region of a gray scale image, i.e., about mean.

3) Many algorithms suffer from the fact that there is no natural method to

determine the number of optimum thresholds. After applying the present algorithm recursively a few times, PSNR of the thresholded image is found to saturate. This property can be used to obtain the appropriate number of thresholds.

## 3  Algorithm

Following steps describe the proposed algorithm for image segmentation:

1. Repeat steps 2-6, $(n\text{-}1)/2$ times; where $n$ is the number of thresholds.

2. Range $\mathbf{R}=[a,\ b]$; initially $a=0$ and $b=255$.

3. Find mean ($\mu$) and standard deviation ($\sigma$) of all the pixels in $\mathbf{R}$.

4. Thresholds $T_1$ and $T_2$ are calculated as $T_1 = \mu - \kappa.\sigma$ and $T_2 = \mu + \kappa.\sigma$; where $\kappa$ is a free parameter.

5. Pixels with intensity values in the interval $[a,\ T_1]$ and $[T_2,\ b]$ are assigned a values equal to the respective weighted means of their values.

6. $a = T_1 + 1,\ b = T_2 - 1$.

7. Finally, repeat steps 2-5 with $a = T_1 + 1,\ b = \mu$ and with $a = \mu + 1,\ b = T_2 - 1$.

The number of thresholds $n$ can be chosen depending on the application under consideration. Optimum value of $n$ for an image can be found out using the PSNR vs. $n$ plot of the image as the rate of increase in the PSNR decreases with $n$ and tends to saturate.

The above algorithm thus ensures that the whole image is segmented effectively based on different thresholds found at each stage using simple parameters like mean and standard deviation. Replacing the pixels within a sub-range by a single value leads to enhanced contrast. On the other hand, choosing the weighted mean of a class as the replacement value ensures that intra-class variance of sub-ranges is minimum leading to increased PSNR and quality of image. We also find that for some images, structures can be better extracted if we take the middle value of the sub-range in place of the mean; however this reduces the PSNR value. We see that the sub-range size is large at greater distance from the mean and reduces as we approach the mean. By doing this we are able to zoom in to the region of interest very fast and are able to extract features around mean efficiently. To ensure that a sub-range does not span two different structures or a single structure does not extend beyond a sub-range, the sub-range span is varied by changing the control parameter $\kappa$ at each step. This leads to detailed feature enhancement by preventing clustering of different major structures within a sub-range. Skew in the asymmetric distributions can also be taken care of by using $\kappa_1$ and $\kappa_2$ for thresholds $T_1$ and $T_2$ in each step.

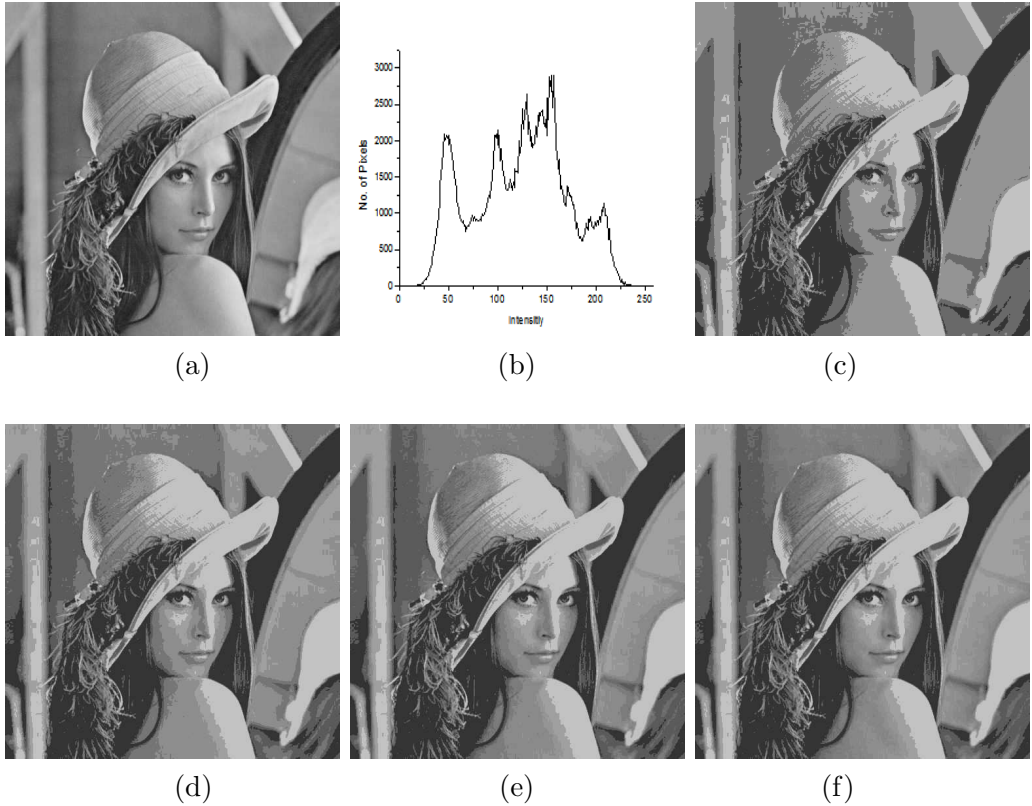

Fig. 1. Results: Lena (a) Lena gray, (b) Histogram, (c) 3 level thresholding, (d) 5 level, (e) 7 level, (f) 9 level

## 4  Results and Observations

For evaluating the performance of the proposed algorithm, we have implemented the method on a wide variety of images. The performance metrics for checking the effectiveness of the method are chosen as computational time so as to get an idea of the complexity, while PSNR is used to determine the quality of the thresholded image. The test images were chosen so as to rigorously test the algorithm for different histogram patterns.

The results of two test images popular in image processing literature are shown in Fig. 1 and Fig. 2. As we increase the number of thresholds the thresholded image rapidly tends to get near to the original image visually. The threshold selection values, computational time and PSNR values for some of the tested images at different number of thresholds are shown in Table 1. For most of the images we observed that (as shown in Fig. 3) the PSNR rapidly saturates after few iterations, providing a criterion for selecting the number of iterations. For simplicity, we have chosen the value of $\kappa$ equal to 1 at all levels of thresholding.

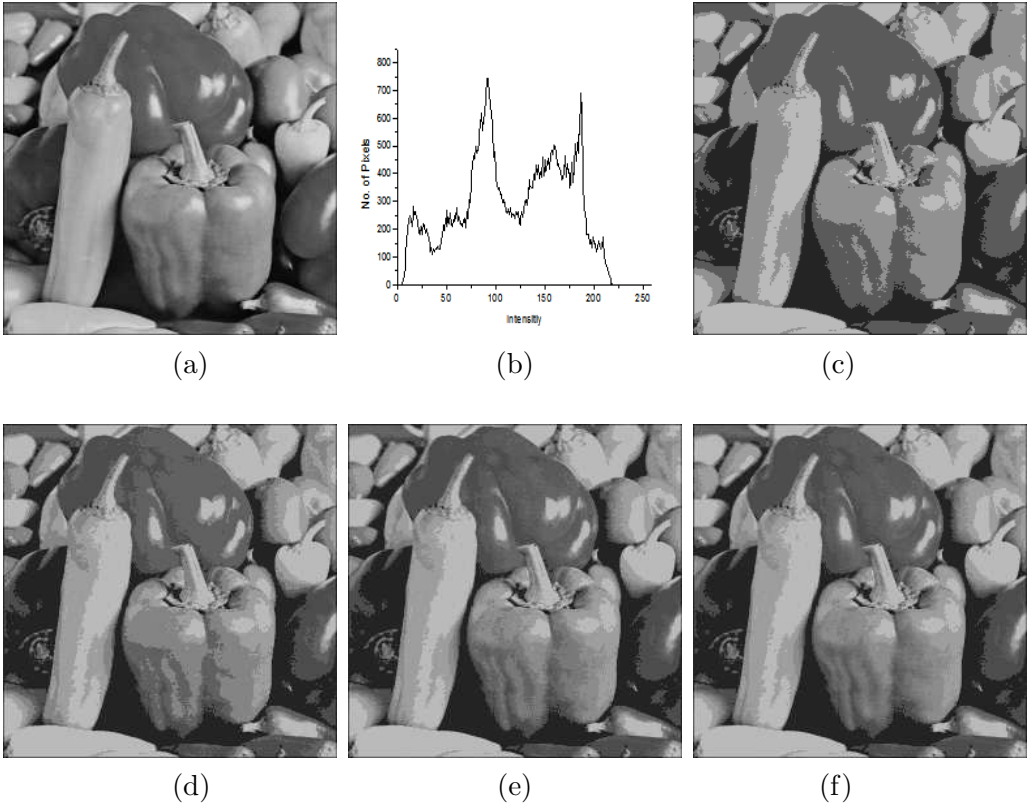

Fig. 2. Results: Peppers (a) Peppers gray, (b) Histogram, (c) 3 level thresholding, (d) 5 level, (e) 7 level, (f) 9 level

It has been observed, Liao et al. (2001), that Otsu's recursive method takes much larger time to calculate multi-thresholds, while our method gives same number of thresholds in relatively less time. From the above results we observe that the algorithm not only segments the image effectively, but also is computationally very fast. We observe that the blocking is non-uniform and the block size reduces near the mean. This gives rise to sharp boundaries and an increase in the contrast.

## 5  Conclusion

A method has been proposed that uses mean and variance of pixel distribution to naturally provide a non-uniform multi-segmentation scheme, ideally suited for human perception. The extreme pixel values are coarse grained in a broader interval as compared to the pixel value distribution around the mean. The procedure naturally adapts to distributions having non-zero higher moments like skew and is quite fast to implement. The recursive procedure converges rapidly as is seen from the quick saturation of the PSNR in variety of images.

| Image | n | Thresholds | Time (milli-sec) | PSNR (dB) |
|---|---|---|---|---|
| Lena (512x512) | 3 | 77, 128, 172 | 203 | 25.84 |
| | 5 | 77, 105, 130, 154, 172 | 281 | 28.56 |
| | 7 | 77, 105, 118, 131, 146, 154, 172 | 359 | 29.33 |
| | 9 | 77, 105, 118, 125, 131, 140, 146, 154, 172 | 437 | 29.51 |
| Baboon (512x512) | 3 | 88, 129, 172 | 204 | 25.99 |
| | 5 | 88, 109, 128, 153, 172 | 297 | 28.41 |
| | 7 | 88, 109, 118, 128, 142, 153, 172 | 391 | 28.97 |
| | 9 | 88, 109, 118, 123, 128, 136, 142, 153, 172 | 469 | 29.08 |
| Peppers (256x256) | 3 | 66, 116, 170 | 33 | 25.25 |
| | 5 | 66, 88, 113, 148, 170 | 50 | 27.53 |
| | 7 | 66, 88, 96, 110, 133, 148, 170 | 62 | 28.29 |
| | 9 | 66, 88, 96, 101, 110, 123, 133, 148, 170 | 79 | 28.49 |
| Jet (512x512) | 3 | 134, 199, 226 | 203 | 24.00 |
| | 5 | 134, 182, 203, 219, 226 | 282 | 25.76 |
| | 7 | 134, 182, 196, 204, 213, 219, 226 | 359 | 26.08 |
| | 9 | 134, 182, 196, 201, 204, 210, 213, 219, 226 | 437 | 26.13 |
| Ariel (256x256) | 3 | 97, 133, 186 | 32 | 25.83 |
| | 5 | 97, 109, 131, 161, 186 | 46 | 28.22 |
| | 7 | 97, 109, 117, 129, 147, 161, 186 | 65 | 28.93 |
| | 9 | 97, 109, 117, 122, 128, 139, 147, 161, 186 | 80 | 29.11 |
| House (256x256) | 3 | 93, 124, 185 | 31 | 24.67 |
| | 5 | 93, 107, 122, 144, 185 | 47 | 27.95 |
| | 7 | 93, 107, 116, 122, 130, 144, 185 | 63 | 28.17 |
| | 9 | 93, 107, 116, 120, 122, 127, 130, 144, 185 | 78 | 28.21 |
| Moon (256x256) | 3 | 102, 130, 156 | 31 | 26.74 |
| | 5 | 102, 117, 130, 145, 156 | 47 | 27.95 |
| | 7 | 102, 117, 124, 130, 139, 145, 156 | 63 | 28.19 |
| | 9 | 102, 117, 124, 128, 131, 136, 139, 145, 156 | 78 | 28.25 |

Table 1
Thresholds and computational time (milliseconds) for the test images.

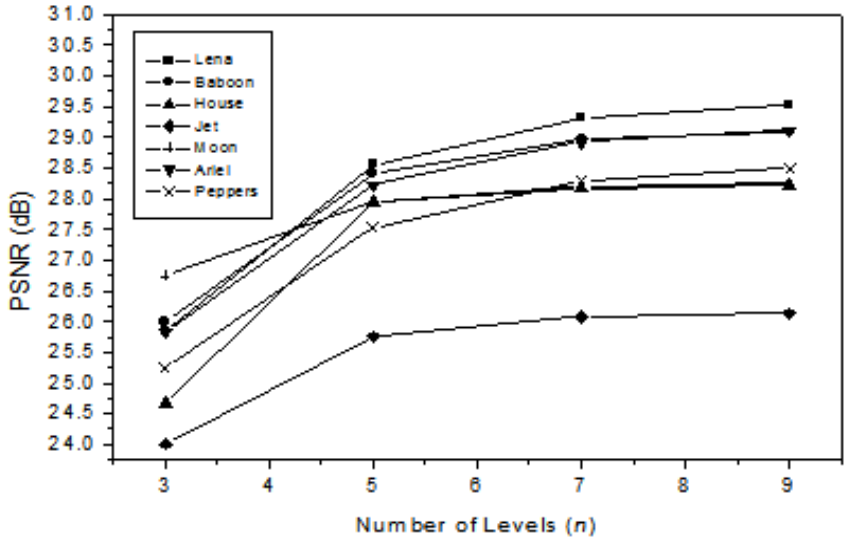

Fig. 3. Plot of PSNR of various test image vs $n$, illustrating saturation of PSNR within a few iterations providing a criterion for optimum number of thresholds.